\documentclass[runningheads]{llncs}
\usepackage{graphicx}
\usepackage{algorithm}
\usepackage{algpseudocode}
\usepackage{subcaption}
\usepackage{tabu}
\usepackage{booktabs}
\usepackage[numbers]{natbib}

\usepackage{tikz}
\usepackage{comment}
\usepackage{amsmath,amssymb} 
\usepackage{color}

\usepackage[capitalize]{cleveref}
\crefname{section}{Sec.}{Secs.}
\Crefname{section}{Section}{Sections}
\Crefname{table}{Table}{Tables}
\crefname{table}{Tab.}{Tabs.}

\newcommand{\figref}[1]{Fig.~\ref{#1}}

\newcommand{\tableref}[1]{Table~\ref{#1}}

\newcommand{\sectionref}[1]{Section~\ref{#1}}
\newcommand{\jc}[1]{\textcolor{black}{#1}}
\newcommand{\jctwo}[1]{\textcolor{black}{#1}}

\newcommand{\jceccv}[1]{\textcolor{black}{#1}}
\newcommand{\jceccvtwo}[1]{\textcolor{black}{#1}}

\usepackage[accsupp]{axessibility}  

\begin{document}
\pagestyle{headings}
\mainmatter
\def\ECCVSubNumber{5340}  

\title{PT4AL: Using Self-Supervised Pretext Tasks for Active Learning} 


\titlerunning{PT4AL: Using Self-Supervised Pretext Tasks for Active Learning}
%
\author{John Seon Keun Yi\inst{1}\thanks{Equal contribution} \and
Minseok Seo\inst{2,4}\protect\footnotemark[1] \and
Jongchan Park\inst{3} \and \\
Dong-Geol Choi\inst{4}\thanks{Corresponding author (dgchoi@hanbat.ac.kr)}}
\authorrunning{JSK. Yi, M. Seo et al.}
%

\institute{Georgia Institute of Technology\inst{1},
SI Analytics\inst{2},\\
Lunit Inc\inst{3},
Hanbat National University\inst{4}}

\maketitle

\begin{abstract}
Labeling a large set of data is expensive. Active learning aims to tackle this problem by asking to annotate only the most informative data from the unlabeled set. We propose a novel active learning approach that utilizes self-supervised pretext tasks and a unique data sampler to select data that are both difficult and representative. We discover that the loss of a simple self-supervised pretext task, such as rotation prediction, is closely correlated to the downstream task loss. Before the active learning iterations, the pretext task learner is trained on the unlabeled set, and the unlabeled data are sorted and split into batches by their pretext task losses. In each active learning iteration, the main task model is used to sample the most uncertain data in a batch to be annotated. We evaluate our method on various image classification and segmentation benchmarks and achieve compelling performances on CIFAR10, Caltech-101, ImageNet, and Cityscapes. We further show that our method performs well on imbalanced datasets, and can be an effective solution to the cold-start problem where active learning performance is affected by the randomly sampled initial labeled set. Code is available at \url{https://github.com/johnsk95/PT4AL}
\end{abstract}

\section{Introduction}
The recent success in deep learning has shown remarkable advancements in computer vision tasks such as classification~\cite{he2016deep,deng2009imagenet} and semantic segmentation~\cite{chen2017deeplab,long2015fully}. 
This has been possible due to the advent of deep convolutional neural networks (CNNs) and large annotated datasets such as ImageNet~\cite{deng2009imagenet} and COCO~\cite{lin2014microsoft}.

\jctwo{As deep learning models are trained in a data-driven manner, having a large enough training set is crucial to achieve high performance.}
However, building a large labeled dataset is prohibitively time-consuming and expensive. 
Labeling costs increase with the size of data and complexity of the tasks.
Instead of labeling the entire data, active learning (AL)~\cite{settles2009active} aims to select \jctwo{informative subsets to label} that achieve the highest performance within a fixed labeling budget. 

\begin{figure}[t]
    \centering
    \includegraphics[width=0.8\columnwidth]{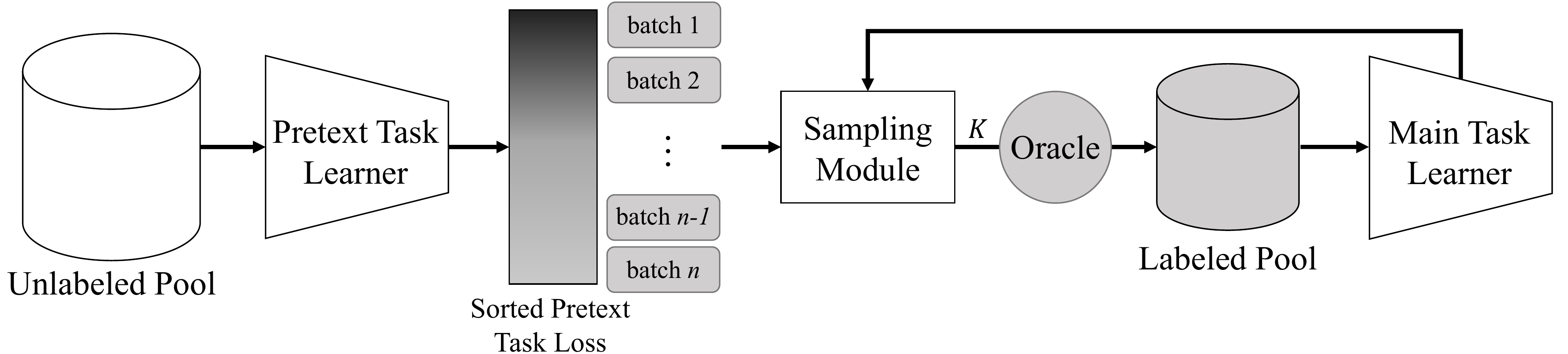}
    \caption{\jctwo{The overall framework of the proposed method. Unlabeled data are sorted by pretext task losses, split into batches, and sampled for training}}
    \label{fig:architecture}
\end{figure}

Existing AL approaches can be divided into two main groups: distribution-based and uncertainty-based methods. Distribution-based methods~\cite{sener2017active,caramalau2021sequential} aim to sample data that well covers the distribution of the feature space. The advantage of such methods is that they can sample \textit{representative} points: data points from high density regions that well represent the overall feature distribution. However, distribution-based sampling fails to select data that are placed near the decision boundary (i.e. high uncertainty data points). 
Uncertainty-based approaches~\cite{lewis1994sequential} resolve this problem by sampling the most uncertain points. 
Simple approaches that utilize class posterior probabilities~\cite{lewis1994sequential,lewis1994heterogeneous}, entropy~\cite{shannon1948mathematical,joshi2009multi}, and loss prediction~\cite{yoo2019learning,kim2021task} were revealed to perform well on various settings. 
While these approaches effectively sample \textit{uncertain} or \textit{difficult} data near decision boundaries in the feature space, they do not capture the overall distribution of the data, according to our qualitative analysis in \figref{fig:embeddings}.
Our method aims to capture the best of both worlds by sampling both representative and difficult data. 

This paper proposes Pretext Tasks for Active Learning (PT4AL), a novel active learning framework that utilizes self-supervised pretext tasks combined with an uncertainty-based sampler. 
We train a pretext task model~\cite{gidaris2018unsupervised,zhang2016colorful} with unlabeled data, and the pretext task loss is highly correlated to the main task loss.
In order to sample diversely from both representative \jceccv{and difficult} data, the unlabeled data are sorted in descending order by their pretext task loss, and split into batches to be used for each AL iteration. 
Starting from the batch containing data with the highest losses, the most uncertain K data points are sampled from each batch, based on the posterior class probability of the previous main task learner.
The uncertainty-based sampler enables PT4AL to sample difficult data, while the batch split allows \jceccvtwo{balanced} sampling across the entire data distribution. 

PT4AL also resolves the innate problem in active learning: the cold start problem. Existing approaches start from a randomly sampled set of labeled data, rendering the overall performance highly dependent on the distribution of the initial set. 
Since our method learns the representation of the unlabeled set in advance, we can sample informative data from the first iteration. 
This approach avoids the issue of high variance and decrease in performance that can stem from randomly sampling the initial labeled set. 

We validate our proposed method on various image classification and semantic segmentation datasets and achieve state-of-the-art or compelling results across different datasets and tasks. Additionally, we demonstrate the robustness of PT4AL on a class imbalanced setting by evaluating on an artificially created class-imbalanced CIFAR10 dataset.

\section{Related Work}
\paragraph{Active Learning} 
Various AL approaches has been proposed, such as information theoretical approaches~\cite{mackay1992information}, ensemble approaches~\cite{mccallumzy1998employing,freund1997selective}, uncertainty based methods~\cite{tong2001support,joshi2009multi} and Bayesian AL methods~\cite{kapoor2007active}. However, these traditional methods have not been verified in large-scale datasets for large-scale models, such as in the field of CNN-based deep learning, which has achieved state-of-the-art in various computer vision tasks.

\jc{Recent AL methods have been centered on large-scale settings for CNN-based deep learning models.}
%
Sener \& Savarese~\cite{sener2017active} proposed a core-set selection method, which chooses data points that cover all data with high diversity based on the feature distribution. This method targets two problems of the previous uncertainty-based methods. First, uncertainty-based methods select only hard samples, resulting in redundant, overlapping data points. Second, the existing methods are not suitable for batch processing on CNNs. The core-set algorithm aims to sample diverse data points in a batch manner.
Yoo \& Kweon~\cite{yoo2019learning} proposed a sub-task module to predict the main task loss of unlabeled data, and sample the high-loss samples from the unlabeled pool.
This method samples from a subset of the unlabeled pool to avoid selecting redundant data points when sampling consecutively from the most uncertain data~\cite{birodkar2019semantic}. 
However, in our qualitative analysis in~\figref{fig:embeddings}, uncertainty-based methods like Yoo \& Kweon sample data points from decision boundaries with less diversity in distribution.
Recently, using a variational autoencoder architecture~\cite{sinha2019variational}, the discriminator adversarially trains the input data to be unlabeled or labeled. In the data sampling phase, a method that first labels the sample predicted as unlabeled with the lowest confidence was proposed.

%

Our active learning method uses a self-supervised pretext task to supplement the flaws of the data distribution-based method and the uncertainty-based method. As described above, AL is largely divided into data distribution-based methods~\cite{sener2017active,caramalau2021sequential,liu2021influence} and uncertainty-based methods~\cite{sinha2019variational,kim2021task,yoo2019learning,cho2022mcdal}. 
The data distribution-based method has the disadvantage that it cannot extract hard samples, and the uncertainty-based method has the possibility to sample overlapping data points and it is difficult to extract the representation of the entire data distribution. Other works~\cite{huang2010active,yang2018variance,behpour2019active} sample from both representative and difficult data by utilizing variance maximization between labeled and unlabeled data or using separate sampling criteria for data in each category. 
Our method uses pretext task-based batch split which allows us to select representative samples across the semantic distribution, and an uncertainty-based in-batch sampler which allows us to select difficult samples.

\paragraph{Representation Learning with Pretext Tasks}
Representation learning aims to learn good pre-trained weights by learning self-supervised pretext tasks with unlabeled data. The pre-trained weights are fine-tuned with a small amount of labeled data to achieve high performance on downstream tasks.
%
The key assumption and the findings in representation learning is that pretext tasks provide enough learning signals without any labels (i.e. direct supervision) provided.
%
Using these assumptions, Liu \textit{et al.}~\cite{liu2020labels} proposed unsupervised neural architecture search (NAS) using self-supervised pretext tasks~\cite{zhang2016colorful,noroozi2016unsupervised,gidaris2018unsupervised} and achieved similar performance to supervised NAS baselines.
%
Zhang \textit{et al.}~\cite{zhang2016colorful} proposed a pretext task to restore the color of the original image through a network after transforming the input image to gray scale.
Noroozi \& Favaro~\cite{noroozi2016unsupervised} improved the performance of representation learning in image classification through the task of dividing input images into grids, mixing them with each other, and inputting each grid into the network.
Gidaris \textit{et al.}~\cite{gidaris2018unsupervised} proposed a pretext task that rotates the input image by $0^\circ$, $90^\circ$, $180^\circ$, and $270^\circ$ and training the network to match the rotated angle of the transformed input image. This method achieved the highest performance among representation learning methods utilizing data structures.
Recently proposed representation learning methods use contrastive learning~\cite{oord2018representation,chen2020simple,chen2021exploring,caron2020unsupervised,he2020momentum} to minimize the distance between different pairwise augmentations of the same image, and repel from augmentations of different images. Contrastive learning is proved to be robust on different downstream tasks and provide state-of-the-art results by far. 
%
%

There have been several efforts to use self-supervised pretext tasks in active learning. Zhu \textit{et al.}~\cite{zhu2020contrastive} uses graph contrastive learning~\cite{you2020graph} for active learning on graph neural networks. \cite{bengar2021reducing,pourahmadi2021simple,hacohen2022active} utilizes self-supervised learning to pre-train the main task model, which is then fine-tuned on labeled data. Bhatnagar \textit{et al.}~\cite{bhatnagar2020pal} presents a multi-task active learner trained for both pretext task and main task, while being robust to mislabeled samples. Although these methods help justify the use of pretext tasks in active learning, they are limited to specific domains~\cite{zhu2020contrastive,pourahmadi2021simple,hacohen2022active}, fail to sample both difficult and representative data~\cite{bengar2021reducing}, and does not solve the cold start problem~\cite{zhu2020contrastive,bengar2021reducing,bhatnagar2020pal}.

As pretext tasks provide good initializations for downstream tasks, we assume that the information learned through these tasks is highly correlated to the semantic data distribution. We analyze and identify the correlation between the pretext task loss and the supervised loss in downstream tasks in Section~\ref{sec:pretext}. Finally, we propose an active learning method using pretext tasks in Section~\ref{sec:method}.

\section{Using Pretext Tasks for Active Learning}\label{sec:pretext}

\jceccv{The success of representation learning with self-supervised pretext tasks~\cite{chen2020simple,he2020momentum,chen2021exploring,liu2020labels}, leads us to believe that there is a high correlation between self-supervised pretext tasks and downstream tasks, and thus pretext tasks can be utilized for active learning.}
\jceccv{Rather than utilizing the feature distribution after the pretext task training, we resort to a simpler metric for active learning - \textit{the pretext task loss}. In this section, we propose and validate a hypothesis, and use these evidences to formulate our AL algorithm. Our hypothesis is that:}
\begin{quote}
    \textit{\textbf{H1}: Pretext task loss is correlated with the main task loss.}
\end{quote}
We think that if a pretext task is correlated or representative of the main task, images that are \textit{hard} (i.e. having high loss values) for the pretext task will also be \textit{hard} for the main task. 
\jceccv{}

\begin{figure*}[t]
    \begin{subfigure}{.33\textwidth}
        \centering
        \includegraphics[width=0.9\columnwidth]{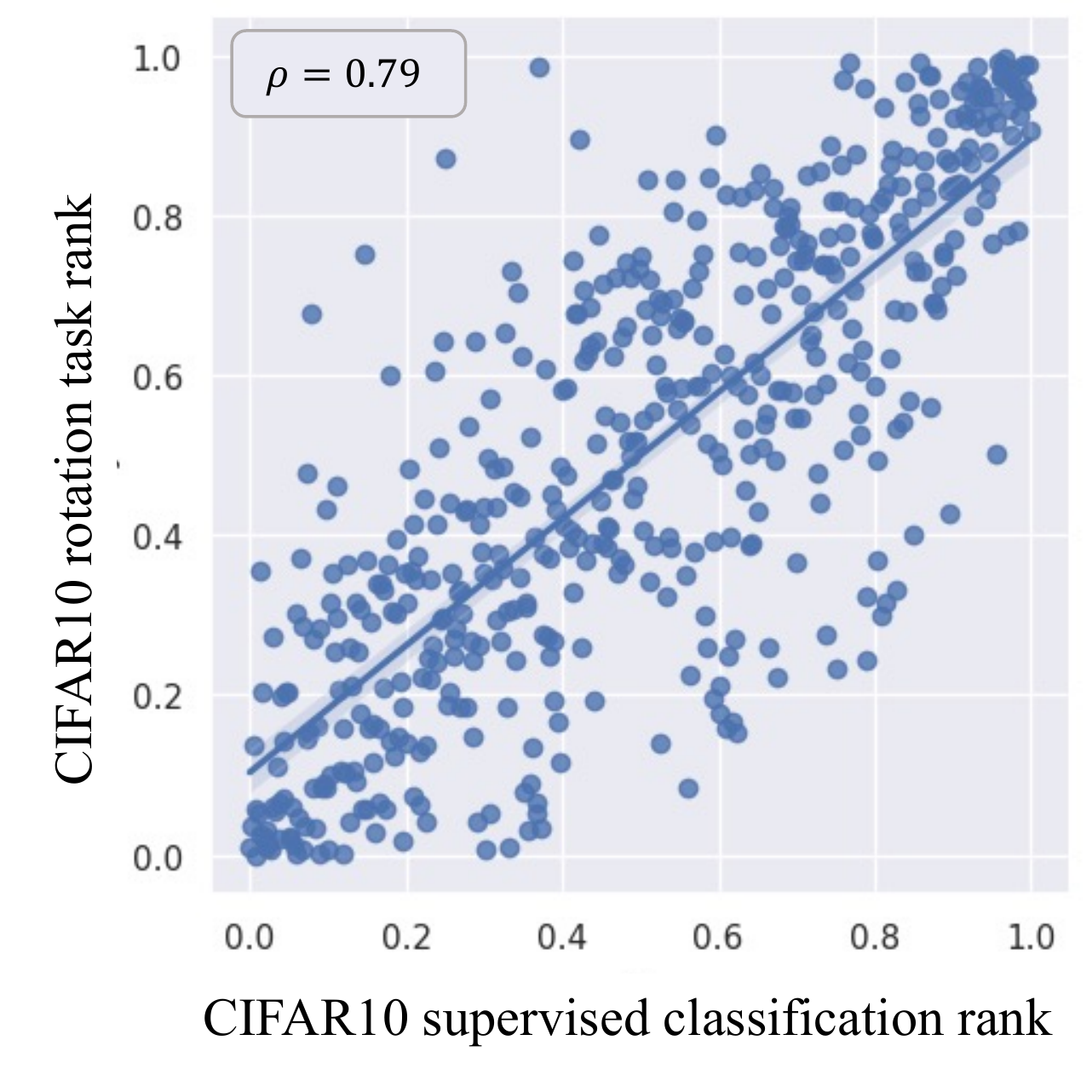}
        \label{fig:cifar_corr}
    \end{subfigure}%
    \begin{subfigure}{.33\textwidth}
        \centering
        \includegraphics[width=0.9\columnwidth]{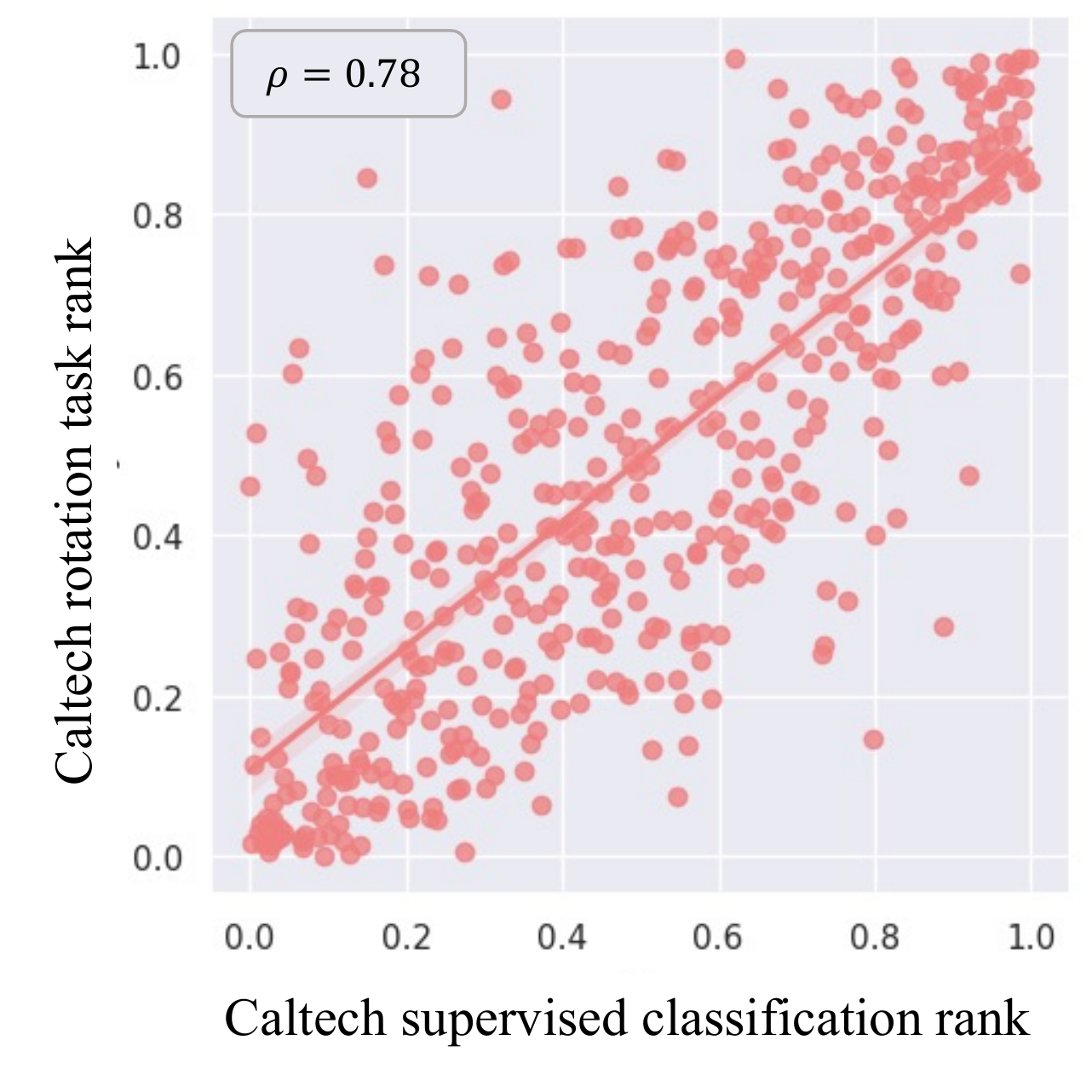}
        \label{fig:caltech_corr}
    \end{subfigure}%
    \begin{subfigure}{.33\textwidth}
        \centering
        \includegraphics[width=0.9\columnwidth]{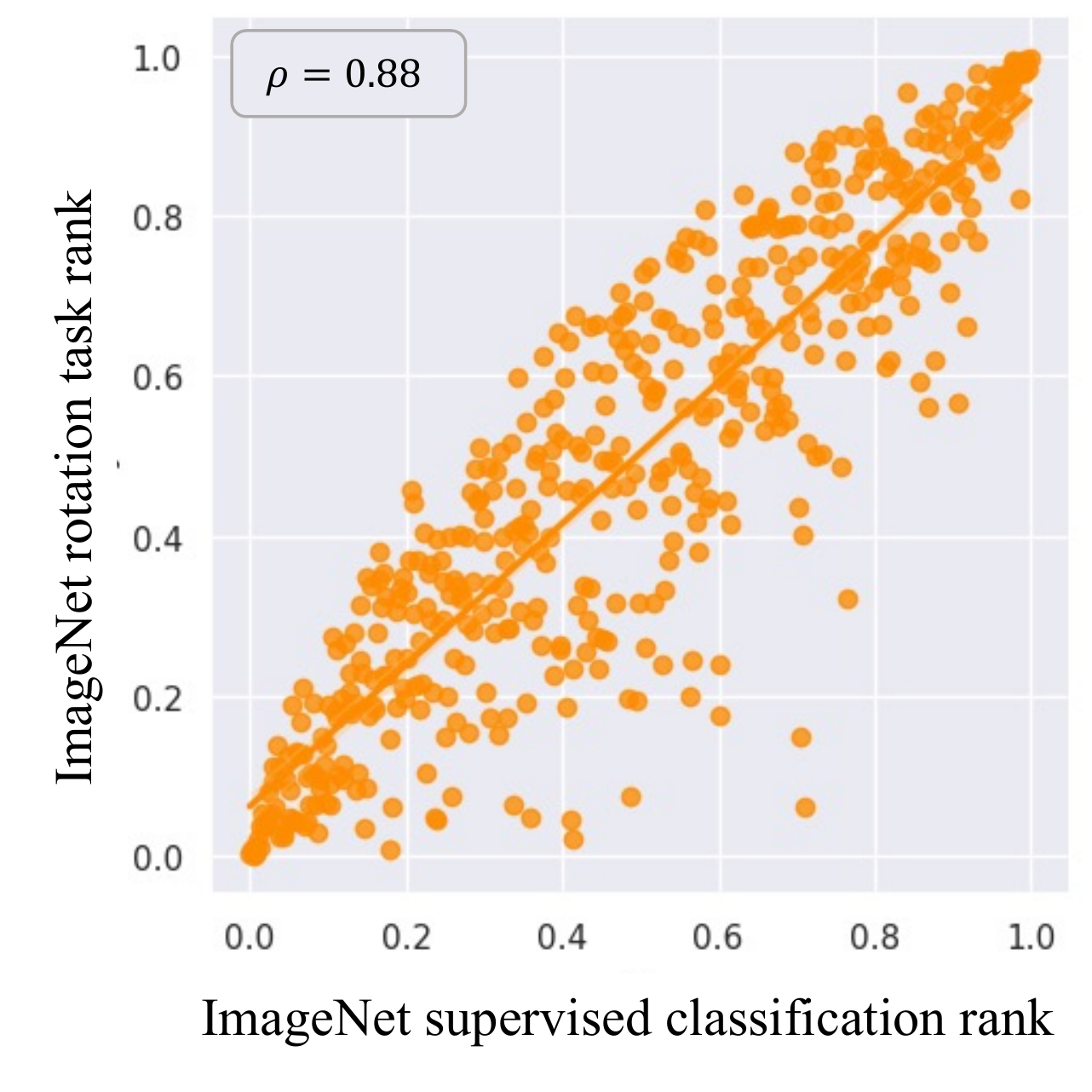}
        \label{fig:imagenet_corr}
    \end{subfigure}
    \caption{(From left to right) The loss rank correlation plots for the main task loss and the pretext task loss in CIFAR10, Caltech-101 and ImageNet. The x and y axes represent the normalized rank of the two losses, respectively}
    \label{fig:correlation}
\end{figure*}

\jceccv{\figref{fig:correlation} presents scatter plots of the pretext task loss and the main task loss in three benchmark datasets. The x-axis is the normalized rank of the main task loss, and the y-axis is the normalized rank of the pretext task loss. The pretext task and the main task are independently trained with the training set, and the losses are computed in the test set. For ease of interpretation, we visualized 1,000 random samples on the plots. Spearman's rank correlation~\cite{spearman1961proof} denoted as $\rho$ is calculated on the full test set.}

As illustrated in~\figref{fig:correlation}, the pretext and main task losses have a strong positive correlation. That is, if a data sample has high loss for a pretext task, it is likely for it to have high loss for the main task, and vice versa. We observe high $\rho$ values for all three datasets: CIFAR10 ($\rho=0.79$), Caltech-101 ($\rho=0.78$), and ImageNet ($\rho=0.88$). Note that these datasets vary in image size, number of classes, and class balance. 
\jceccv{The strong correlation between the pretext task loss and the main task loss across diverse datasets validates our hypothesis, and thus is a strong evidence for using pretext task losses for active learning.}
%
However, there is one caveat to the hypotheses: methods that use contrastive loss as the pretext task~\cite{he2020momentum,chen2020simple,chen2021exploring} do not have a strong loss correlation. ($\rho=-0.001$ for SimSiam) We contribute this result to two main reasons: class bias of the contrastive loss and strong reliance to augmentations. Details are explained in the supplementary material. Even if we could find a way to achieve close correspondence with the main loss, we decide not to use contrastive methods since the large batch size and long training time generally required for these methods beat our purpose of a simple and quick AL model. Details are explained in the supplementary material.

\jceccv{Throughout this work, we validate the efficacy of PT4AL with 4 different pretext tasks: Rotation prediction~\cite{gidaris2018unsupervised}, colorization~\cite{zhang2016colorful}, solving jigsaw puzzles~\cite{noroozi2016unsupervised}, and SimSiam~\cite{chen2021exploring}.}
\jceccv{We compare and analyze the efficacy of different pretext tasks on classification and semantic segmentation in \sectionref{sec:ablation_pretext}.}
\jceccv{Since rotation prediction~\cite{gidaris2018unsupervised} performs best in CIFAR10 and colorization~\cite{zhang2016colorful} performs the best in Cityscapes, we use rotation prediction for image classification main tasks, and colorization for semantic segmentation.}

\section{Method}\label{sec:method}
In this section, we introduce the specifics of PT4AL. First, we provide a brief overview of our active learning algorithm. Then we provide details of the pretext task learning for batch split and in-batch sampling in the following sections. 

\subsection{Overview}\label{sec:overview}
\jceccv{In a typical active learning scenario, we are initially provided with a pool of unlabeled data $x_U \in X_U$. The objective of AL is to achieve the best performance in the main task model $F_m(\cdot)$ with a limited amount of labeled data. 
In specific, we follow the \textit{batch mode} active learning scheme: in the $i$-th AL iteration, we select $K$ samples from $X_U^i$, add them into labeled pool $(X_L^i, Y_L^i)$ with oracle, train and evaluate $F_m^i(\cdot)$ with $(X_L^i, Y_L^i)$. The iterations are repeated until the specified labeling budget is reached.}

\jceccv{The overall framework of PT4AL is illustrated in~\figref{fig:architecture}. PT4AL is split into two parts: pretext task learning for batch split and in-batch sampling. Pretext task learning is done prior to the AL iterations. We train a pretext task learner with $X_U$. The unlabeled samples are sorted in descending order of their pretext task losses, and split into batches. The in-batch sampling is done at each AL iteration. At the $i$-th iteration, the sampling module selects $K$ samples from the $i$-th batch, according to the uncertainty of the main task learner in these samples. The main task learner $F_m^i(\cdot)$ is trained with $(X_L^i, Y_L^i)$ and evaluated on the test set.}


\subsection{Pretext Task Learning for Batch Split}\label{sec:pretext_learning}
\jceccv{In this section, we explain how a pretext task is used for active learning batch split. The term \textit{batch} refers to a pool of unlabeled data to be sampled in an AL iteration. While any pretext task can be used in our method, we use the widely used rotation prediction task~\cite{gidaris2018unsupervised} for the explanation.}
For the rotation prediction task, the backbone neural network~\cite{he2016deep} is trained on all four orientations ($0^\circ$, $90^\circ$, $180^\circ$, $270^\circ$ degrees) of the input image. The loss function is defined as the average of the losses for each orientation:
\begin{equation}
  loss(x_i,\theta_p) = \frac{1}{k}\sum_{y=1}^k\mathcal{L}_{CE}(F_p(g(x_i\mid y)\mid \theta_p), y)
  \label{eq:rotation_loss}
\end{equation}
Where $\mathcal{L}_{CE}$ is the cross-entropy loss. The rotation operator $g(\cdot\mid y)$ yields the rotated input image according to the orientation label $y$. We define $k=4$ since we predict four different rotations. $F_p$ represents the probability distribution of the input image rotated by label y. Note that the rotation label $y$ is unknown to $F_p$. In inference, four orientations of each image is fed into the trained network $F_p$ and the extracted loss is the same averaged loss $loss(x_i,\theta_p)$ used in training. $F_p$ is trained and tested on the same unlabeled set $X_U$. The model weights $\theta_p$ with the best test accuracy is used for loss extraction.

\jceccv{After training the pretext task learner, we extract pretext task loss values from $X_U$ and split them into batches. Given the pretext task loss values of the unlabeled data $loss_{X_U}$ in the pretext task learning phase, we first sort the losses in descending order. The sorted data $\mathcal{X}_U$ is then divided into $I$ batches of equal size. The number of $I$ is equal to the number of AL iterations: if there are ten iterations($I=10$), there will be ten batches $\mathcal{B}=\{b_i\}_{i=1}^{I=10}$.}

\subsection{In-batch Sampling}\label{sec:data_sampling}

\jceccv{The in-batch sampler selects $K$ samples at each AL iteration. 
At the $i$-th iteration, the in-batch sampler $\phi(\cdot)$ selects $K$ samples from the $i$-th batch to be annotated by the oracle. 
The sampler computes the top-1 posterior probability in the given batch using the previous main task learner $F_m^{i-1}$, and $K$ data points with the lowest confidence scores are selected. 
In the first iteration, $K$ points are sampled from the first batch $b_0$ at even intervals.
Equation \ref{eq:k_sampler} summarizes the sampler $\phi(\cdot)$.
The sampling makes use of the main task model from the previous iteration, $F_m^{i-1}$. }

\begin{equation}
  \phi(b_i, F_m^{i-1}) = min_K\{max(F_m^{i-1}(b_i \mid \theta_m))\}
  \label{eq:k_sampler}
\end{equation}

\noindent \jceccv{Algorithm~\ref{alg:sampling} illustrates our overall sampling algorithm including batch splitting and in-batch sampling. In the first iteration when we do not have $F_m^0$, we uniformly select samples in the first batch, based on our empirical observation that visually similar samples have similar pretext task loss values.}
\begin{algorithm}
    \caption{Sampling Strategy}\label{euclid}
    \begin{algorithmic}
        \State \textbf{Input:} Unlabeled $X_U$, labeled $X_L$, pretext task losses $loss_{X_U}$, main task model $F_m$
        \State $\mathcal{X}_U = sort(loss_{X_U})$ \Comment{Sort losses in descending order}
        \State Split $\mathcal{X}_U$ into batches $\mathcal{B}$
        \For {$b_i$ in $\mathcal{B}$}
            \State if i == 1, $X_K = uniform(b_i, loss_{X_U}^i)$ \Comment{\jceccv{For the first batch, uniformly sample}}
            \State else, $X_K = \phi(b_i, F_m^{i-1})$ \Comment{For other batches, sample top-K uncertain data}
            \State $X_U \gets X_U - X_k$ \Comment{Remove from unlabeled pool}
            \State $X_L \gets X_L \cup X_k$ \Comment{Add to labeled pool}
            \State train $F_m^i$ with $X_L$
        \EndFor
    \end{algorithmic}
    \label{alg:sampling}
\end{algorithm}
Sampled data have two main traits: difficult and representative. Difficult or uncertain data refers to data that the main task model cannot easily distinguish because it is near a decision boundary. Conversely, representative data well defines the distribution in the feature space. 
Our intuition is that if we can sample data from both categories, we can form a labeled pool with the most informative data. \jc{This is empirically verified through query analysis in Section~\ref{sec:experiments}.} Our batch split method combined with the sampler samples both representative and difficult data. Our method is much simpler and well performing compared to previous works that sample data from both traits~\cite{huang2010active,yang2018variance,behpour2019active}.


\section{Experiments}\label{sec:experiments}
We evaluate the efficacy of our method on two commonly used visual recognition tasks: image classification and semantic segmentation.
We choose CIFAR10~\cite{krizhevsky2009learning}, Caltech-101~\cite{fei2004learning}, ImageNet~\cite{deng2009imagenet} benchmarks for image classification, and Cityscapes~\cite{cordts2016cityscapes} for semantic segmentation.
To further demonstrate our method's efficacy in a more challenging class-imbalanced setting, we additionally use a class-imbalanced version of CIFAR10. Finally, we show the use of PT4AL as an effective solution to the cold start problem. Unless otherwise specified, all the experiment results are reproduced by ourselves, averaged over multiple runs with different random seeds.


\subsection{Image Classification}\label{sec:img}
\paragraph{Dataset} We perform experiments on three image classification datasets with varying size and number of classes. CIFAR10 contains 50,000 training and 10,000 testing images of size 32 $\times$ 32 with 10 object categories. We start with 1,000 labeled images, and 1,000 images are added for each iteration. Caltech-101 has 9,144 images of size around 300 $\times$ 200 distributed around 101 classes. We divide the data into 8,046 for training and 1,098 for testing. Similar to CIFAR10 we also start with 1,000 labeled images with increments of 1,000 per iteration. ImageNet consists of over 1.3M images of 1,000 classes. 1,279,867 and 49,950 images are used for the training and testing set. For ease of experimentation and to avoid noise from similar class labels, ImageNet classes are reduced to 67 based on the WordNet~\cite{pedersen2004wordnet} superclasses. ImageNet starts with $K \approx 128,000$ labeled samples, and the same $K$ samples are selected for each iteration. Due to heavy computation, each ImageNet performance is the average of 3 runs. 
\paragraph{Baselines and implementation details} We compare PT4AL with random sampling, Core-Set~\cite{sener2017active}, Variational Adversarial Active Learning (VAAL)~\cite{sinha2019variational}, Learning Loss~\cite{yoo2019learning}, CoreGCN~\cite{caramalau2021sequential}, and PAL~\cite{bhatnagar2020pal}.   
For CIFAR10 we add ``Learning loss(detached)", \jc{where the loss prediction task is detached during supervised learning to avoid influences from multi-task learning.} ResNet-18~\cite{he2016deep} is used as the backbone network for the pretext task and the main task learner. The final linear layer of the pretext task learner is converted to (512,4) to account for the four orientations of the rotation task. For Caltech-101 and ImageNet, input images are resized into 224 $\times$ 224. No data augmentation is applied in the pretext task learning phase. Random resized crop and horizontal flip is applied in the main task phase. The main task is trained for 200 epochs in CIFAR10 and Caltech-101, and 100 epochs in ImageNet. SGD with a multi-stage learning rate is applied. Detailed hyper-parameters are described in the supplement material.
\begin{figure*}[t]
    \begin{subfigure}{.33\textwidth}
        \centering
        \includegraphics[width=1.\columnwidth]{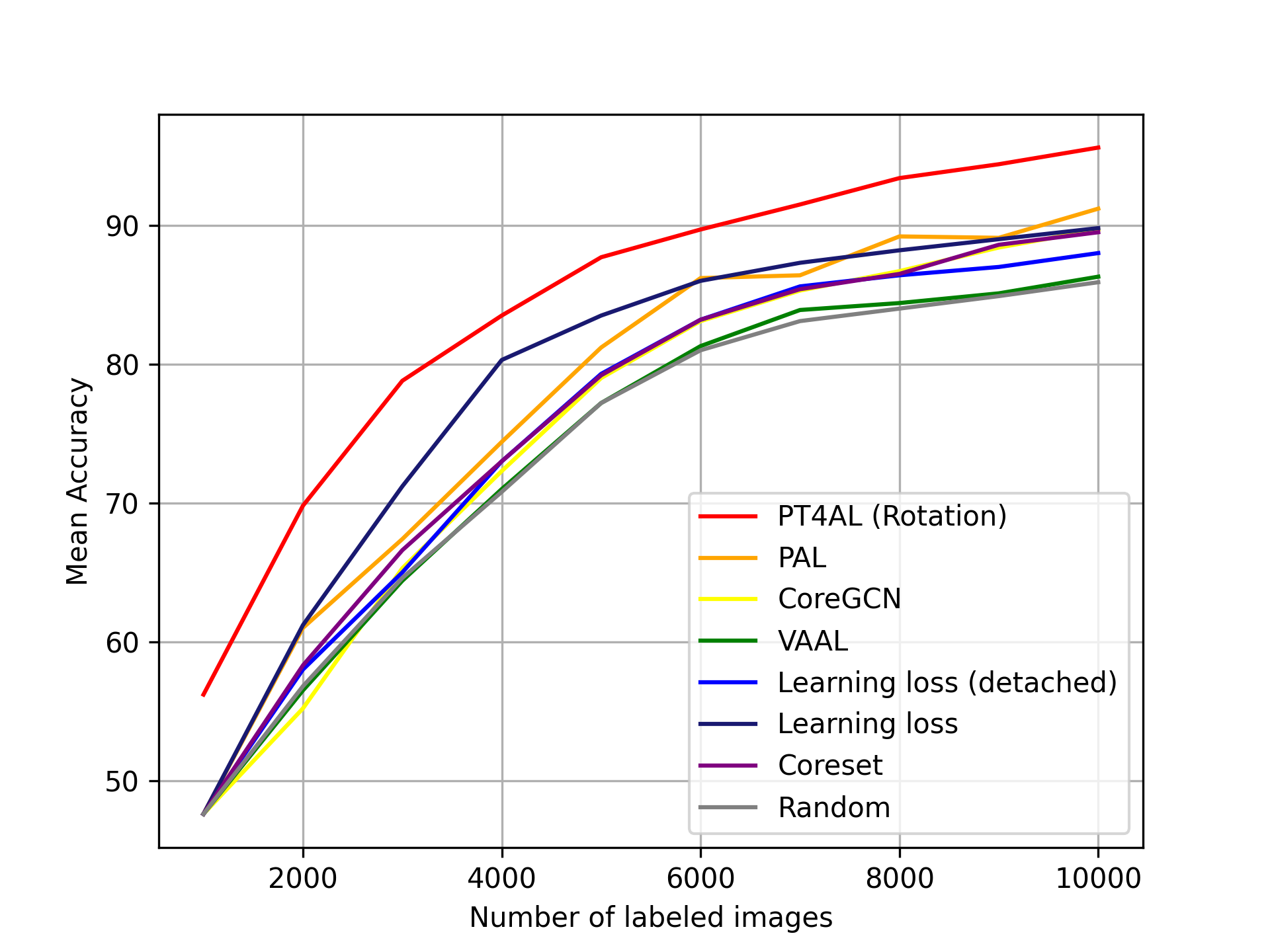}
        \caption{CIFAR10}
        \label{fig:cifar_perf}
    \end{subfigure}%
    \begin{subfigure}{.33\textwidth}
        \centering
        \includegraphics[width=1.\columnwidth]{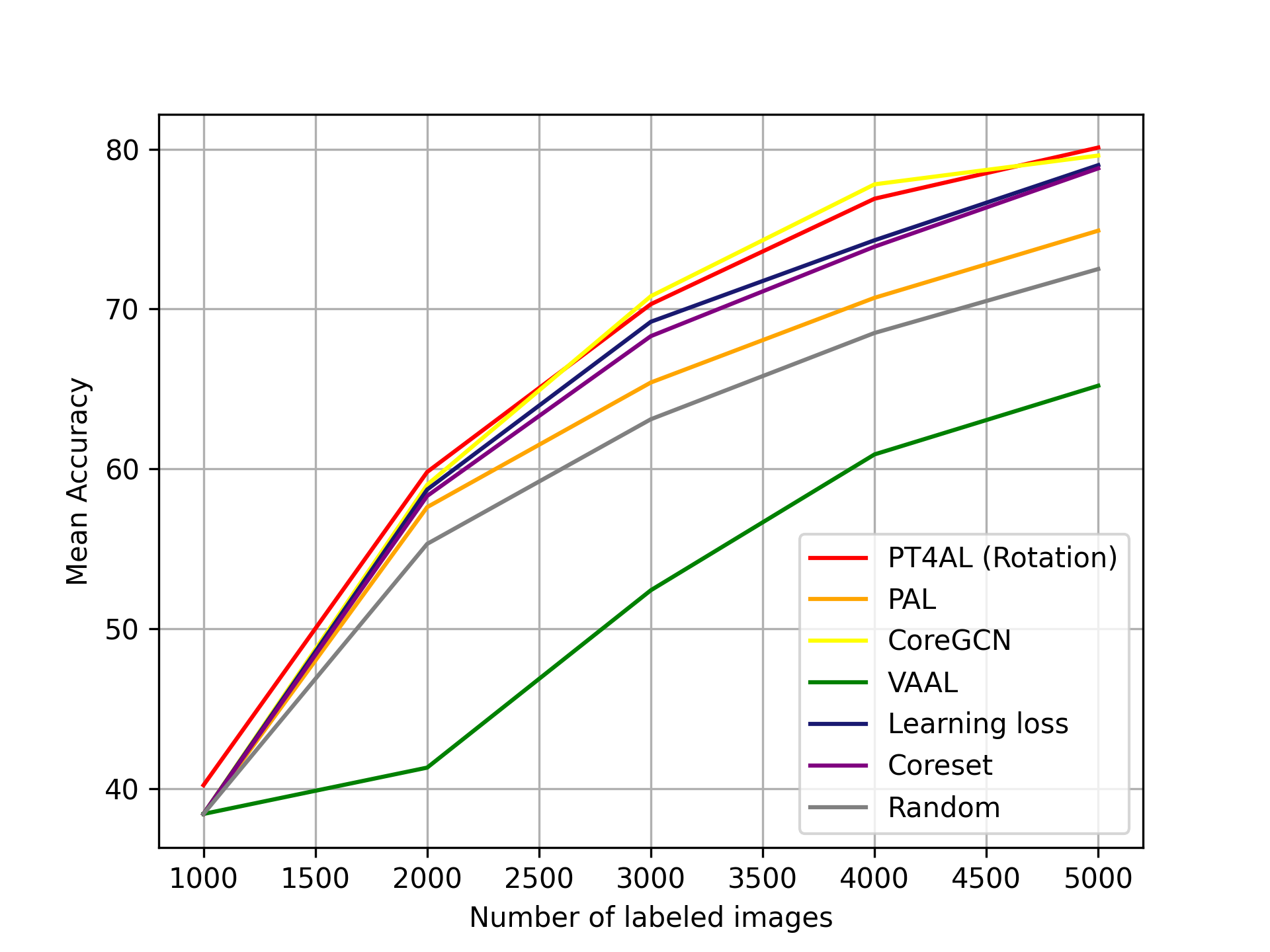}
        \caption{Caltech-101}
        \label{fig:caltech_perf}
    \end{subfigure}%
    \begin{subfigure}{.33\textwidth}
        \centering
        \includegraphics[width=1.\columnwidth]{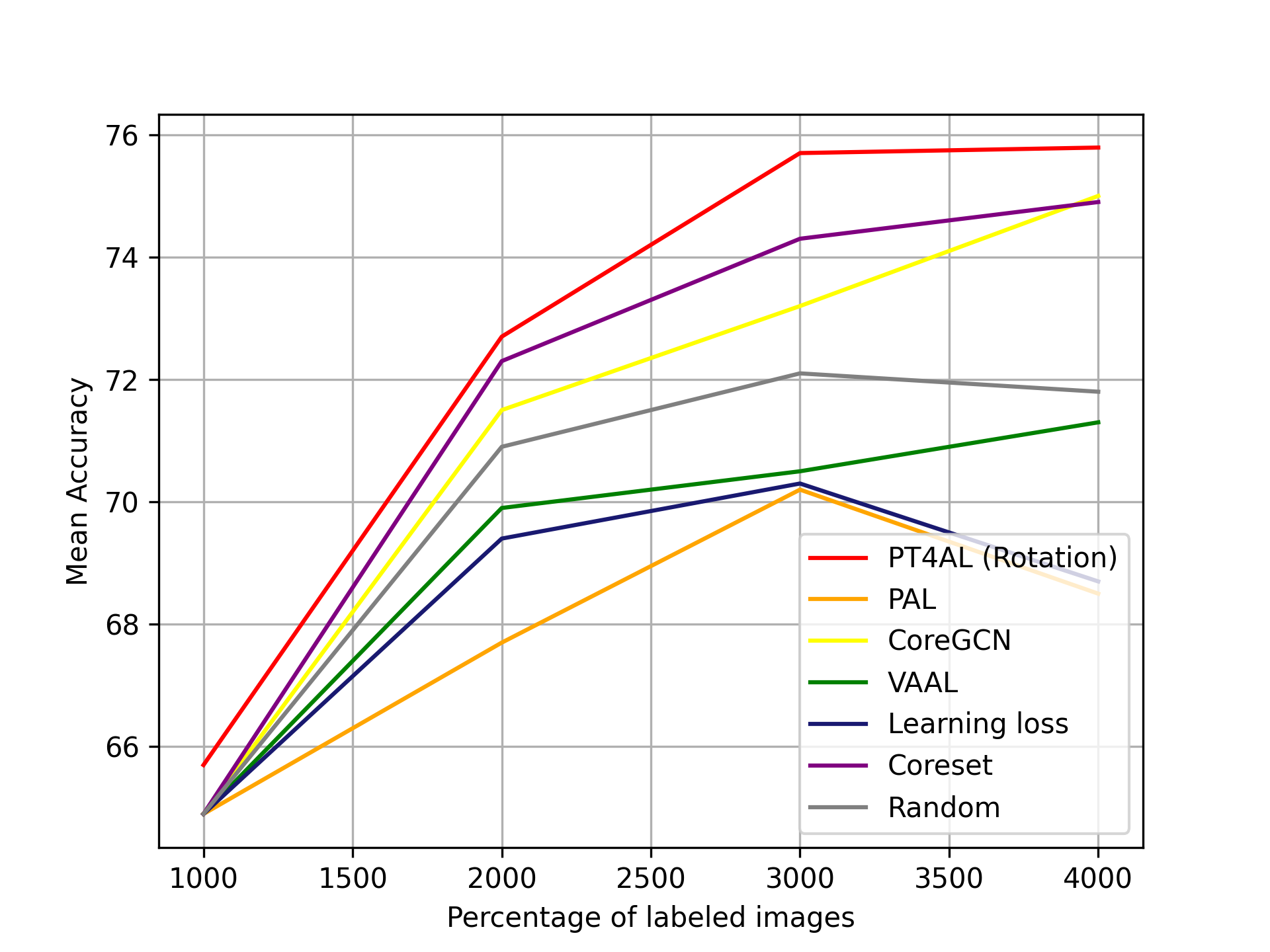}
        \caption{ImageNet-67}
        \label{fig:imagenet_perf}
    \end{subfigure}
    \caption{Comparison of image classification performance on CIFAR10, Caltech-101, ImageNet-67. Best viewed in color}
    \label{fig:classification_results}
\end{figure*}

\paragraph{Results}
~\figref{fig:cifar_perf} demonstrates the results for CIFAR10.
\jceccv{PT4AL clearly outperforms other methods across all AL iterations by a noticeable margin.}
The accuracy of PT4AL in the final iteration of 10,000 labeled points is 95.13\% ($\uparrow$ 8.91\%), while the second-best performing learning loss scores 89.93\% ($\uparrow$ 3.71\%). 
Note that detached learning loss~\cite{yoo2019learning} performs significantly worse than the original multi-task learning approach, \jceccv{where the main task model is simultaneously trained with auxiliary tasks}. 
\jceccv{The significant drop in performance due to the detachment indicates that the multi-task approaches~\cite{yoo2019learning,sinha2019variational,caramalau2021sequential} may benefit from multi-task learning. To strictly measure the benefit of AL to select informative samples, we need to compare the detached setting across all methods.}
Our method also has a significant advantage from the first iteration, achieving an accuracy of 55.83\% ($\uparrow$ 9.81\%) compared to the other methods' 46.02\%. 
\jceccv{This emphasizes the advantage of PT4AL sampling informative points in the first iteration, instead of random sampling in other AL frameworks. Further details of PT4AL solving the cold-start problem is described in \sectionref{sec:cold_start}.}

\jceccv{Similar results can also be observed in Caltech-101 and ImageNet, in \figref{fig:caltech_perf} and \figref{fig:imagenet_perf}. Our method outperforms other methods across most of the iterations with a considerable advantage from the start.}



\begin{figure*}[t]
    \begin{subfigure}{.33\textwidth}
        \centering
        \includegraphics[width=0.8\columnwidth]{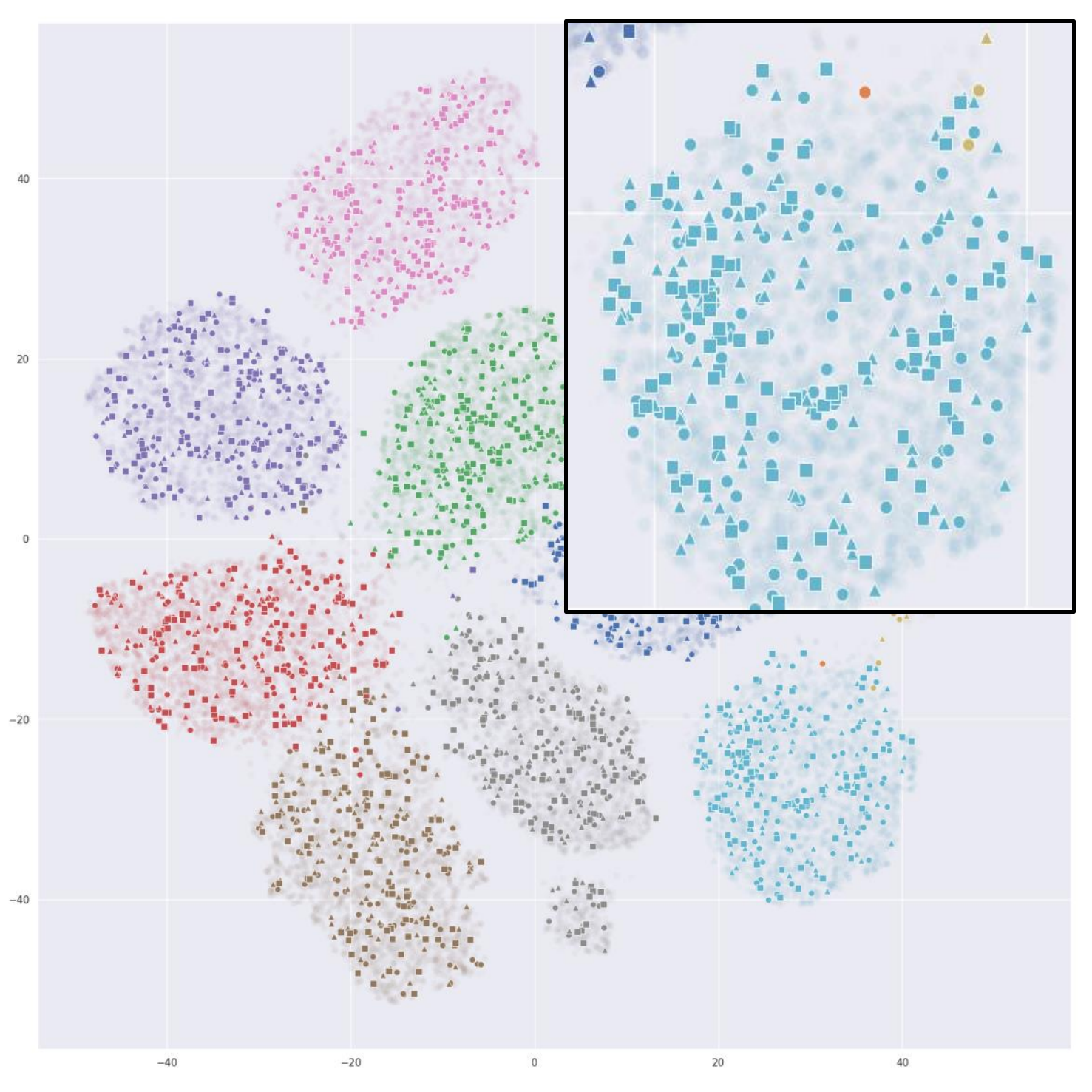}
        \caption{Random}
        \label{fig:tsne1}
    \end{subfigure}%
    \begin{subfigure}{.33\textwidth}
        \centering
        \includegraphics[width=0.8\columnwidth]{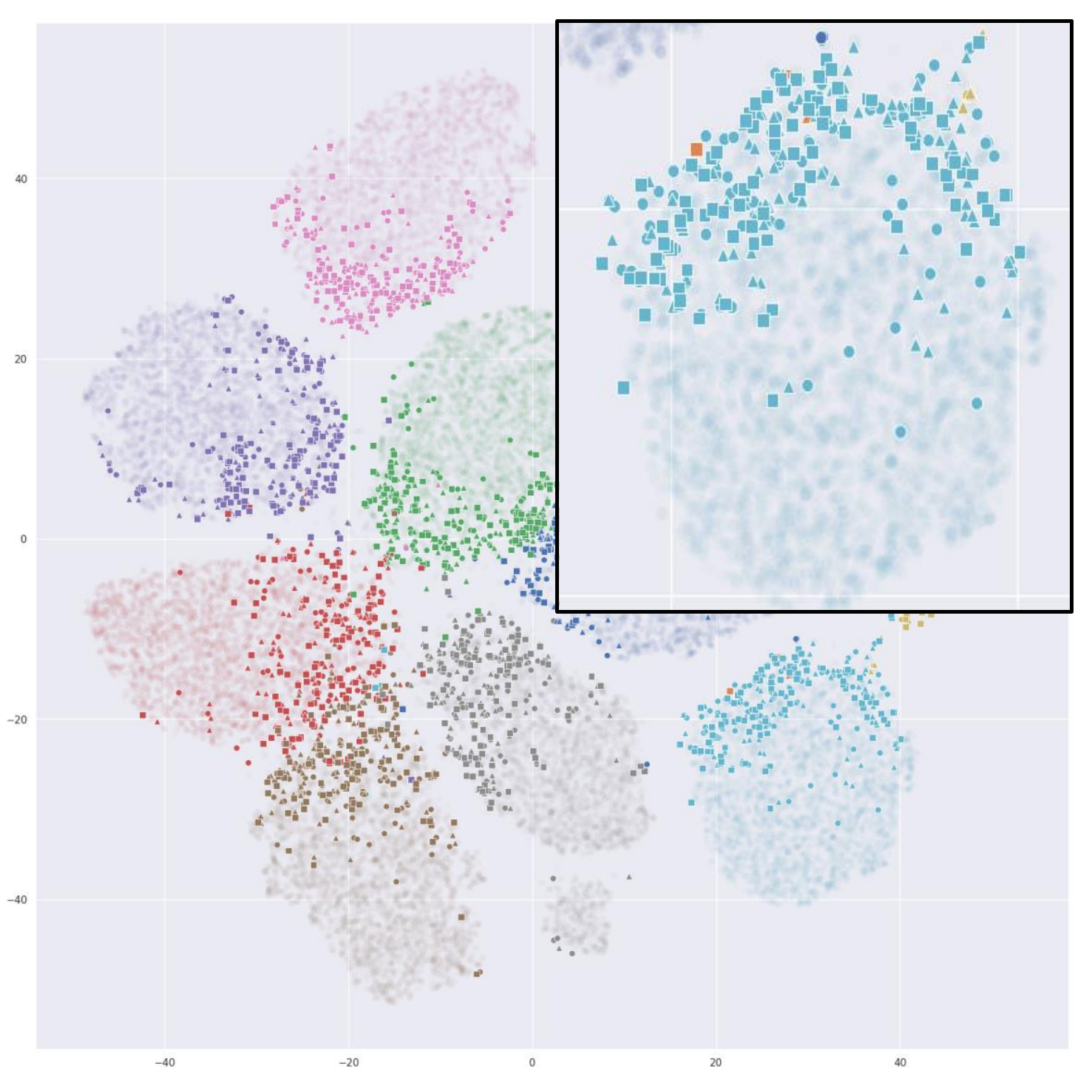}
        \caption{Learning loss}
        \label{fig:tsne2}
    \end{subfigure}%
    \begin{subfigure}{.33\textwidth}
        \centering
        \includegraphics[width=0.8\columnwidth]{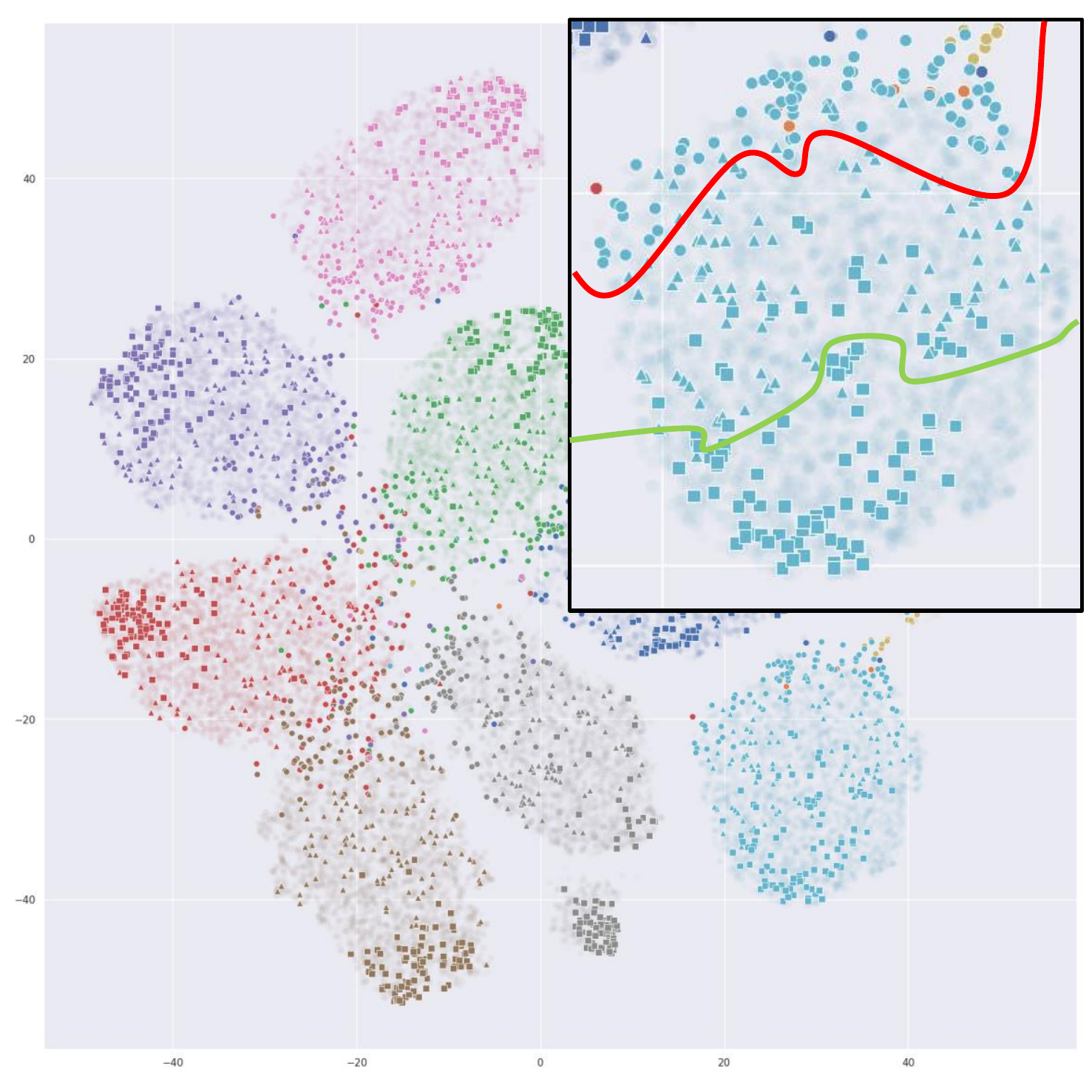}
        \caption{PT4AL(Rotation)}
        \label{fig:tsne_ours}
    \end{subfigure}%
    \caption{t-SNE visualization of the CIFAR10 dataset for random, learning loss~\cite{yoo2019learning} and PT4AL. Vivid points are sampled for labeling. Best viewed in color}
    \label{fig:embeddings}
\end{figure*}

\paragraph{Query Analysis}
~\figref{fig:embeddings} illustrates t-SNE~\cite{van2008visualizing} embeddings of the CIFAR10 data points sampled by random, learning loss~\cite{yoo2019learning} and ours. For a fair comparison, we use embeddings extracted from a ResNet-18 model trained with fully labeled CIFAR10. 
\jceccv{To visualize the sampled data in different methods across the AL iterations, each of the 1,000 samples from the first iteration are marked in circle, fifth iteration in triangle, and tenth (last) iteration as square.}
\figref{fig:tsne1} shows that random sampling queries evenly from the embedding space, but fails to sample difficult data points along the decision boundaries. As shown in \figref{fig:tsne2}, learning loss~\cite{yoo2019learning} has most of its queries concentrated on the border regions. While this may be effective for the labeled classifier to learn difficult points, it does not query points that represent the classes well. \jceccv{\figref{fig:tsne_ours} shows that PT4AL queries from both difficult and representative regions. The sampled points are either concentrated on the class boundaries or evenly located in the class distributions.}
Since PT4AL initially samples from batches with higher pretext loss values, selected points from the first iteration are concentrated on the decision boundaries of the embedding space. As the sampler progresses to batches with lower loss values, we can see that the sampled points propagate to the remaining regions of the class clusters. Such sampling behavior is a mix of both distribution and uncertainty-based methods, mitigating their flaws while sampling both difficult and representative data points. 

\subsection{Semantic Segmentation}\label{sec:sem_seg}

\begin{figure*}[t]
    \centering
    \begin{subfigure}{.44\textwidth}
        \centering
        \includegraphics[width=1.\columnwidth]{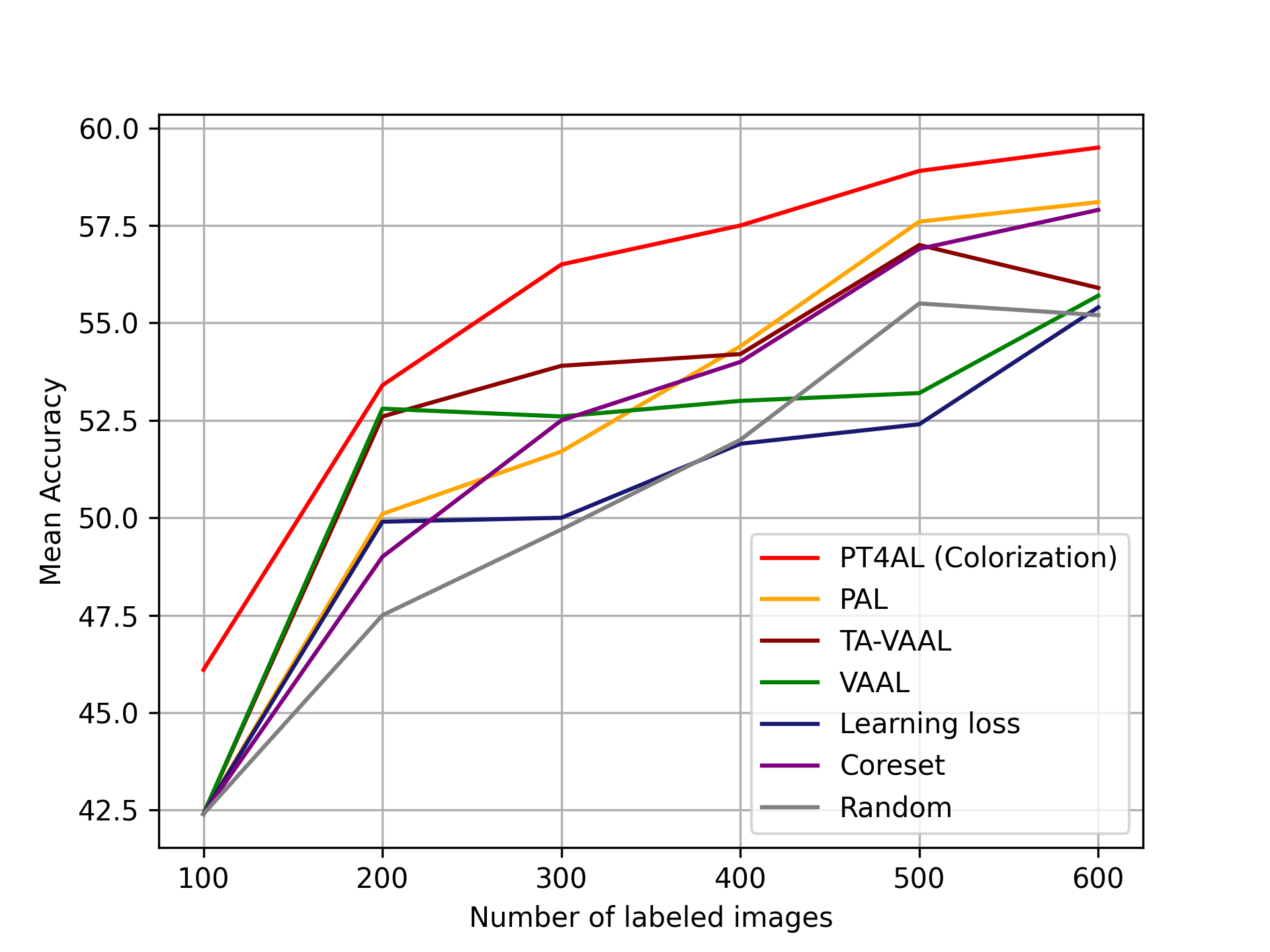}
        \caption{Cityscapes}
        \label{fig:city}
    \end{subfigure}%
    \begin{subfigure}{.44\textwidth}
        \centering
        \includegraphics[width=1.\columnwidth]{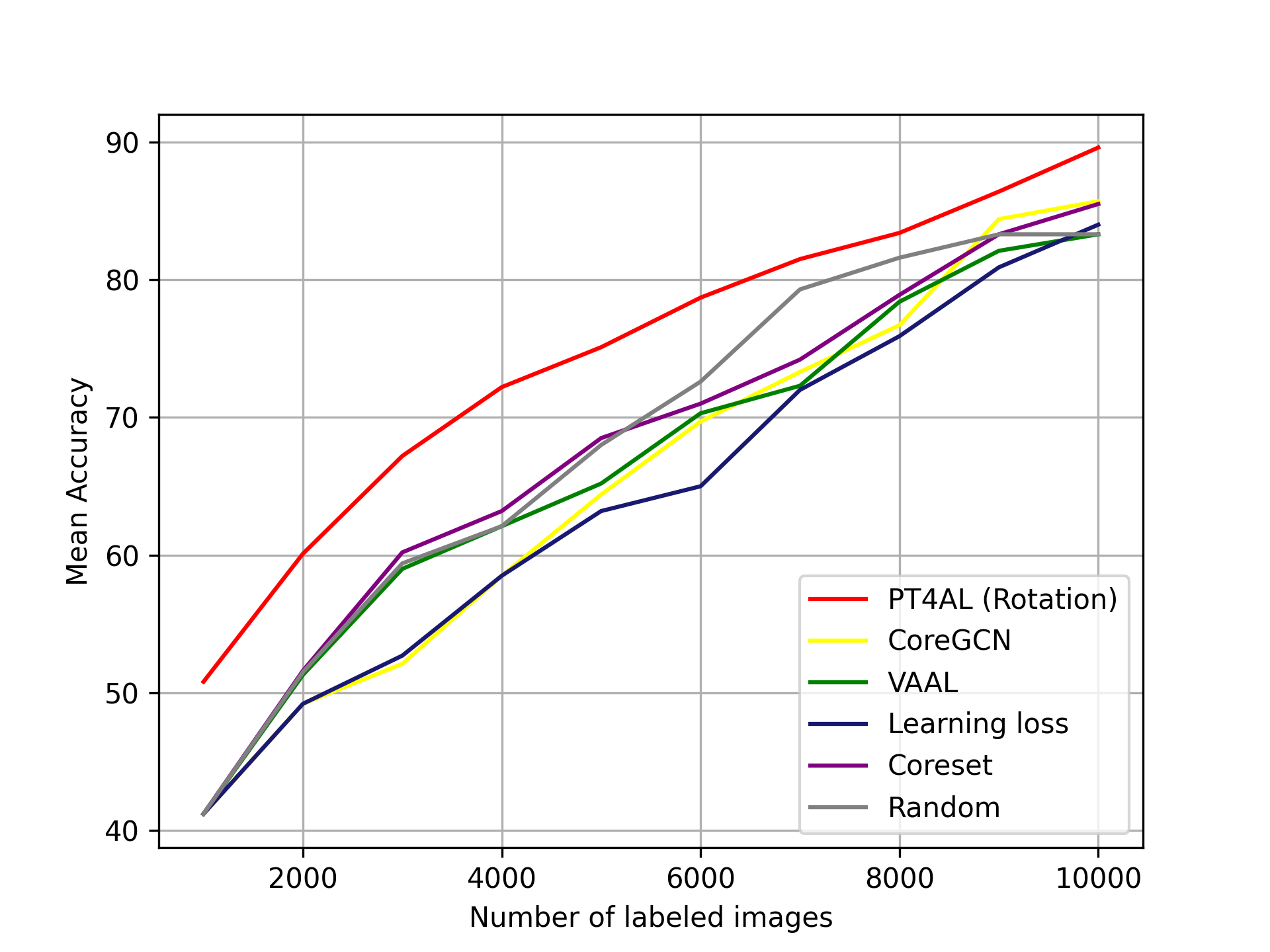}
        \caption{Imbalanced CIFAR10}
        \label{fig:im_cifar}
    \end{subfigure}
    \caption{Comparison of cityscapes semantic segmentation and imbalanced cifar10 classification. Best viewed in color}
    \label{fig:seg_im_results}
\end{figure*}
\paragraph{Dataset} We choose Cityscapes~\cite{cordts2016cityscapes}, a public benchmark dataset widely used in semantic segmentation. The dataset consists of 2,975 training and 500 validation images. 
\jceccv{At each AL iteration, 100 images are sampled for the labeled training set. The original training set is set as the unlabeled set.}

\paragraph{Baselines and Implementation Details}
We choose the state-of-the-art active learning methods for this experiment: Core-Set~\cite{sener2017active}, Learning loss~\cite{yoo2019learning}, VAAL~\cite{sinha2019variational}, TA-VAAL~\cite{kim2021task}, and PAL~\cite{bhatnagar2020pal}. \jc{We choose a widely-used semantic segmentation architecture, DeepLab~\cite{chen2017deeplab} with a ResNet-101~\cite{he2016deep} backbone. The model is initialized with ImageNet~\cite{5206848} pre-trained weights.}
The input images are resized to (1024,512) and no data augmentation is applied. The training batch size is 1, and all hyper-parameters follow the original paper~\cite{chen2017deeplab}, unless otherwise specified. 

%

\paragraph{Results}
\jceccv{\figref{fig:city} demonstrates that PT4AL outperforms all other methods by a noticeable margin across all iterations. The performance improvement at the first iteration is also significant, showing that PT4AL is an effective solution for the cold start problem. Note that learning loss~\cite{yoo2019learning}, VAAL~\cite{sinha2019variational} and Core-Set~\cite{sener2017active} are not as effective as in CIFAR10, sometimes being worse or on-par with the random selection baseline. One possible reason for the universal effectiveness of PT4AL across tasks is that the nature of PT4AL can dynamically change with respect to the pretext task being used. A more detailed analysis among pretext tasks is described in \sectionref{sec:ablation_pretext}.}

\subsection{Image Classification on an Imbalanced Dataset}\label{sec:imbalanced}
\paragraph{Dataset \& Experiment details} 
To evaluate the efficacy of PT4AL on a more challenging class-imbalanced setting, we recompose the CIFAR10 dataset. The number of images for each class is as follows:  airplane-500, automobile-1,000, bird-1,500, cat-2,000, deer-2,500, dog-3,000, frog-3,500, horse-4,000, ship-4,500 and truck-5,000. All implementation details are identical to the balanced CIFAR10 described in \ref{sec:img}, except for dataset composition.

%
%

\paragraph{Results}
\figref{fig:im_cifar} demonstrates the performances of PT4AL and other baselines on imbalanced CIFAR10.
PT4AL outperforms other baselines across all iterations by a large margin.
%
In the pretext task, data from classes with little training data generally have high loss, and classes that have abundant training data generally have low loss values. 
Since PT4AL samples from data batches with high to low loss, it can sample in a class-balanced way even in imbalanced settings. \jceccv{Also, unlike other methods using only the main task model related metrics, PT4AL utilizes the pretext task loss which is completely independent from the main task model.}
\jceccv{Interestingly}, unlike the experiment results in balanced CIFAR10, data distribution-based AL methods (Core-Set, CoreGCN) obtains higher performance than the uncertainty-based methods (VAAL, learning loss). 
These results empirically show that uncertainty-based methods are more negatively affected by the class-imbalanced setting than the distribution-based methods. PT4AL outperforms other methods by a margin, showing robustness on a more challenging class-imbalanced setting. Furthermore, we observe that PT4AL samples data in a more class-balanced way. Details on the class distribution of the sampled data are in the supplementary material.

\subsection{Cold Start Problem in Active Learning}\label{sec:cold_start}
\jceccv{Since most AL approaches require a trained main task model, the first AL iteration starts with randomly selected labeled data. This is what we call \textit{the cold start problem in active learning}.}
To thoroughly validate the efficacy of our method as a solution to the cold start problem, we take a closer look into the first AL iteration in the CIFAR10 benchmark. Note that all other methods use random selection for the first iteration.
For PT4AL, after training the pretext task learner, the unlabeled data are sorted by pretext task loss in a descending order, split into 10 batches, and 1,000 data points are uniformly selected from the first batch. The experiment is repeated 20 times with different random seeds. All implementation details are identical to \sectionref{sec:img}.

%
\tableref{table:cold} summarizes the experiment results. 
%
PT4AL displays more stable performance compared with random sampling in the first iteration, as the standard deviation is smaller, and the gap between the max/min accuracy is smaller than that of the random baseline.
%
PT4AL significantly outperforms random in the average accuracy, indicating that more informative data points are sampled for the main task model.
These results indicate that PT4AL is a good solution to the cold start problem, and can be used as a good starting point for existing AL methods~\cite{yoo2019learning,sinha2019variational,kim2021task,liu2021influence,caramalau2021sequential}. More details are in the supplement material.
\begin{table}[h!]
\centering
\caption{Results of the first active learning iteration in CIFAR10}
\begin{tabu}{c|c|c}
\tabucline[1pt]{-}
\textbf{Method}& \textbf{Mean accuracy}&\textbf{Min / Max}  \\ \hline
Random & $47.49 \pm 3.15 \%$  & 43.06 / 53.74 \%   \\
PT4AL(Rotation) & $55.20 \pm 1.95$ \% & 52.00 /  57.71 \%   \\ 
\tabucline[1pt]{-}
\end{tabu}
\label{table:cold}
\end{table}

\subsection{Computational Overheads}
\jceccv{As described in \sectionref{sec:method}, the extra computation for PT4AL, apart from the main task model training, is the pretext task learning for batch split, and the unlabeled data inference for uncertainty measurement in in-batch sampling.}
\jceccv{To fairly compare the computational overheads of different approaches, we measure the wall-clock time of the methods compared in CIFAR10 experiment under the same environment. In \figref{fig:cifar_perf} and \tableref{table:computation}, we can observe that PT4AL achieves the best performance while having on-par computational overheads with others. Core-Set~\cite{sener2017active} has similar computations as the random selection baseline.}

\begin{table}[h!]
\centering
\caption{The wall-clock time of each algorithm in the CIFAR10 experiment}
\begin{tabu}{c|c|c|c|c|c}
\tabucline[1pt]{-}
\textbf{Method}& Random Selection& \textbf{PT4AL} & Learning Loss~\cite{yoo2019learning} & VAAL~\cite{sinha2019variational} & CoreGCN~\cite{caramalau2021sequential}  \\ \hline
\textbf{Time} & \jceccv{2hr 16min} & \textbf{3hr 36min} & 2hr 37min & 14hr 9min & 3hr 48min  \\
\tabucline[1pt]{-}
\end{tabu}
\label{table:computation}
\end{table}

\section{Ablation Study}


\begin{figure*}[t]
    \centering
    \begin{subfigure}{.33\textwidth}\label{fig:ablation_component}
        \centering
        \includegraphics[width=1.\columnwidth]{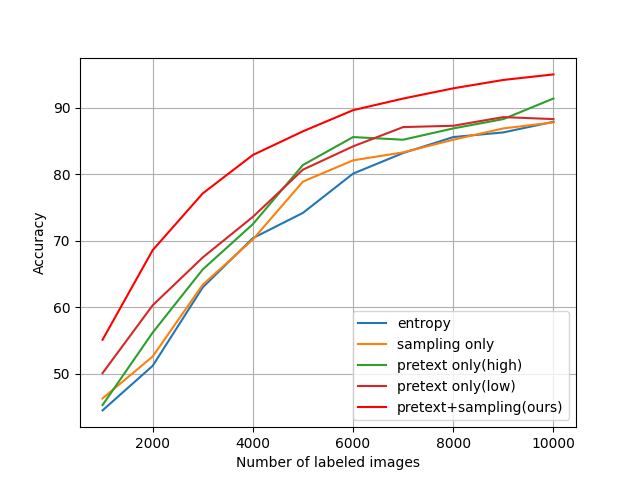}
        \label{fig:ablation_component}
        \caption{Component Ablation}
    \end{subfigure}%
    \begin{subfigure}{.33\textwidth}\label{fig:pretext_cifar10}
        \centering
        \includegraphics[width=1.\columnwidth]{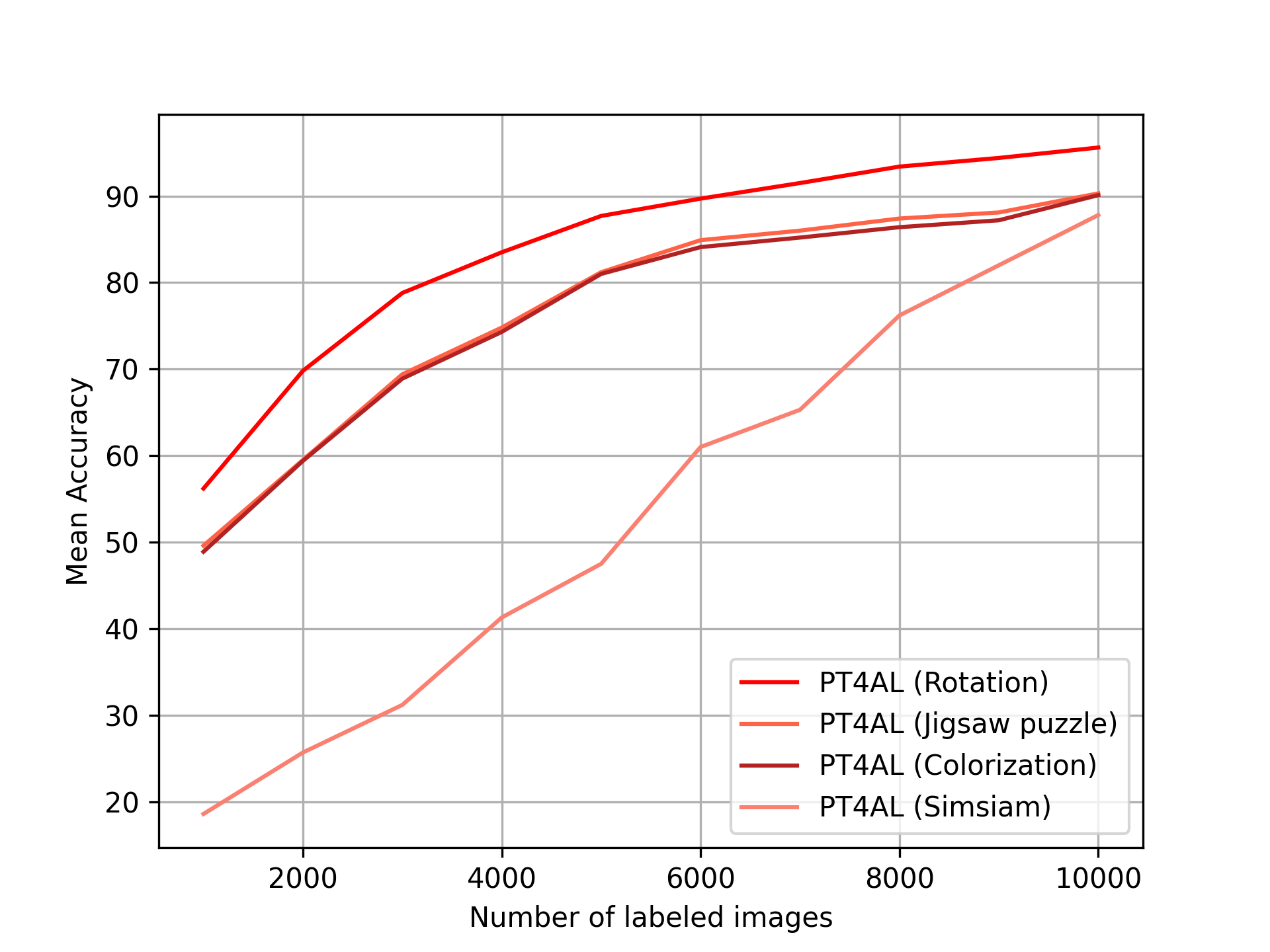}
        \label{fig:ablation_cifar}
        \caption{CIFAR10}
    \end{subfigure}%
    \begin{subfigure}{.33\textwidth}\label{fig:pretext_city}
        \centering
        \includegraphics[width=1.\columnwidth]{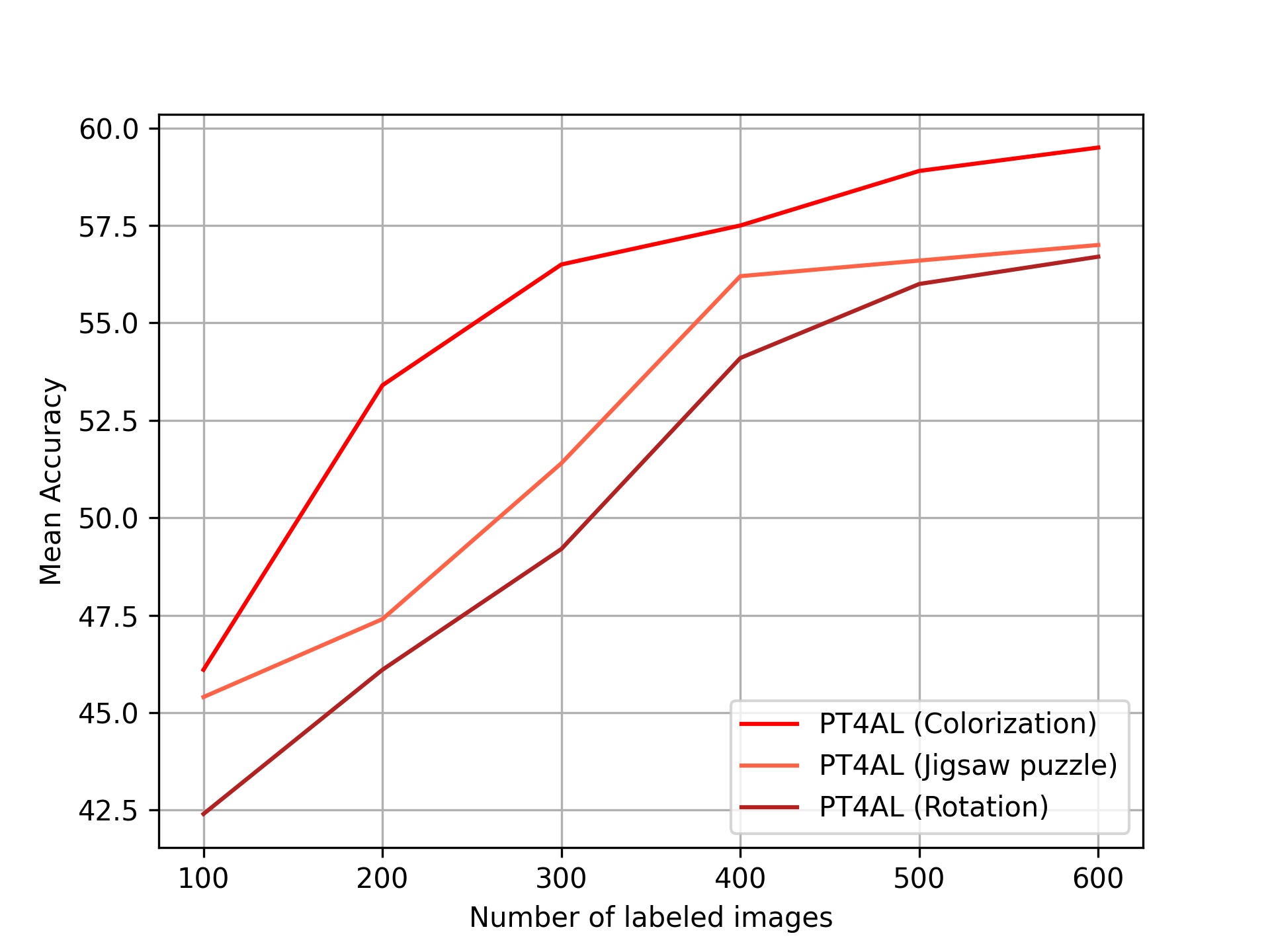}
        \label{fig:ablation_cityscapes}
        \caption{Cityscapes}
    \end{subfigure}%
    \caption{(a) Ablation on two components of PT4AL. (b),(c) PT4AL with different pretext tasks on CIFAR10 and Cityscapes}
    \label{fig:pretext} 
\end{figure*}


\subsection{Ablation on sampling strategy and pretext task loss}
\figref{fig:pretext}.a shows the ablations results of the two core components of PT4AL. Instead of the pretext task loss, ``Sampling only" uses the main task model's entropy. Batches are made by randomly segmenting the unlabeled data. Data in the first iteration are randomly sampled as there is no main task model to begin with. ``Pretext only" replaces our proposed sampling method with a naive sampling of high-loss samples or low-loss samples. Compared with PT4AL which uses both heuristics, the two variations display inferior performance. Using both components is imperative to a well-performing model. 

\subsection{Pretext Tasks}\label{sec:ablation_pretext}
\jceccv{\figref{fig:pretext}.b and \figref{fig:pretext}.c presents active learning performance of PT4AL using different pretext tasks. Rotation prediction~\cite{gidaris2018unsupervised}, colorization~\cite{zhang2016colorful}, solving jigsaw puzzles~\cite{noroozi2016unsupervised}, and SimSiam~\cite{chen2021exploring} are compared in CIFAR10 and Cityscapes benchmarks. The experiment settings are identical to \sectionref{sec:img} and \sectionref{sec:sem_seg}. The inferior performance of SimSiam is analyzed in detail in the supplement material. The rotation prediction task shows the best performance in CIFAR10, and the colorization task performs best in Cityscapes. The best performing pretext task differs by the main task. Since rotation prediction is an image-level task and colorization is a pixel-level task, it is intuitive to match rotation prediction with image classification and colorization with segmentation.}

\subsection{Sampling Strategy}\label{sec:sam}
\paragraph{Sampling in the first iteration}
In the first iteration, we do not have access to the main task model for uncertainty measurement. Thus, sampling within the first batch resorts to sampling with the pretext task losses. We compare three simple sampling methods in CIFAR10: top-K, random, and uniform.
%
%
The performances for top-K loss, random, and uniform sampling are 44.65\%, 51.88\%, and 55.20\%, respectively. 
As the uniform sampling outperforms the other two sampling methods, we choose it as our in-batch sampling method for the first iteration.
%
We observe that the samples with similar loss values are visually similar, indicating overlapping semantic information in the top-K sampling. This observation is also coherent with the best performance of uniform sampling, as it avoids selecting data points with visually too similar data points. More details are in the supplementary material.
%
%
\paragraph{High loss first vs Low loss first batch split} We examine two different strategies for batch split: high loss batch first or low loss batch first. \textit{High loss batch first} method starts the first iteration with the batch containing the highest pretext task losses, then moves to batches with lower losses for consecutive iterations. \textit{Low loss batch first} is in reverse. On the CIFAR10 experiment, the high loss first strategy displays slightly better results with 55.20\% accuracy in the first iteration and 95.13\% from the last iteration. Low loss first strategy scores 53.47\% and 94.59\% in the first and last iterations. 
We attribute the small performance difference between the two batch split methods to a finding in curriculum learning~\cite{wu2020curricula}. Low loss batch first and high loss batch first are analogous to curriculum learning and anti-curriculum learning, respectively. Wu~\textit{et al.}~\cite{wu2020curricula} concludes that curriculum or anti-curriculum is not effective in standard settings, which explains the small performance gap. 
As the high loss first method performs better across all iterations, we choose it as our batch split method.

%

%
\section{Conclusion}
\jceccvtwo{In this paper we introduce PT4AL, a novel active learning method based on pretext tasks. We demonstrate the correlation between pretext tasks and semantic recognition tasks, and utilize the pretext task losses to split unlabeled samples into batches. In the query analysis in \sectionref{sec:experiments}, we show that the batches are scattered across the whole semantic distribution. Combined with the uncertainty-based in-batch sampler, PT4AL samples both difficult and representative data from the unlabeled pool.}
We thoroughly examine our method on two widely used vision tasks across various datasets. Our method demonstrates compelling results on datasets with varying resolution, scale and class distribution. We also show that PT4AL is an effective solution for the cold start problem. Although our proposed method performs well on different tasks and datasets, performance varies by the pretext task being used and some tasks such as SimSiam~\cite{chen2021exploring} perform poorly. 
\jceccvtwo{Future research directions may include designing a pretext task that is universal across various recognition tasks.} \\

\noindent{\textbf{Acknowledgments}}: This research was supported by the the National Research Foundation of Korea(NRF) grant funded by the Korea government(MSIT) (No. 2022R1F1A1075019, No. 2021M3E8A2100446) and partially supported by the IITP grant funded by the MSIT (No. 2014-3-00123).

%
%
\bibliographystyle{splncs04}
\bibliography{egbib}
\end{document}